\documentclass[10pt, conference, letterpaper]{IEEEtran}
\IEEEoverridecommandlockouts
\usepackage{cite}
\usepackage{amsmath,amssymb,amsfonts}
\usepackage{caption}
\usepackage{subcaption}
\usepackage{authblk}

\usepackage{algorithmic}
\usepackage{graphicx}
\usepackage{textcomp}
\usepackage{xcolor}
\usepackage{caption}
\usepackage{subcaption}
\usepackage[export]{adjustbox}
\usepackage[noabbrev]{cleveref}
\def\BibTeX{{\rm B\kern-.05em{\sc i\kern-.025em b}\kern-.08em
    T\kern-.1667em\lower.7ex\hbox{E}\kern-.125emX}}

\usepackage{url}
\def\url#1{}

\newtheorem{thm}{Theorem}
\newtheorem{lem}[thm]{Lemma}

\newtheorem{defn}{Definition}[section]

\DeclareMathAlphabet\mathbfcal{OMS}{cmsy}{b}{n}

\newcommand{\mycomment}[1]{}

\DeclareMathOperator*{\argmax}{arg\,max}

\begin{document}

\title{A Novel Switch-Type Policy Network for Resource Allocation Problems: Technical Report \\
\thanks{DISTRIBUTION STATEMENT A. Approved for public release. Distribution is unlimited.

This material is based upon work supported by the Department of the Air Force under Air Force Contract No. FA8702-15-D-0001. Any opinions, findings, conclusions or recommendations expressed in this material are those of the author(s) and do not necessarily reflect the views of the Department of the Air Force.}
}

\author[1]{Jerrod Wigmore}
\author[2]{Brooke Shrader}
\author[1]{Eytan Modiano}
\affil[1]{Massachusets Institute of Technology}
\affil[2]{MIT Lincoln Laboratory}

\maketitle
\begin{abstract}
Deep Reinforcement Learning (DRL) has become a powerful tool for developing control policies in queueing networks, but the common use of Multi-layer Perceptron (MLP) neural networks in these applications has significant drawbacks. MLP architectures, while versatile, often suffer from poor sample efficiency and a tendency to overfit training environments, leading to suboptimal performance on new, unseen networks. In response to these issues, we introduce a switch-type neural network (STN) architecture designed to improve the efficiency and generalization of DRL policies in queueing networks. The STN leverages structural patterns from traditional non-learning policies, ensuring consistent action choices across similar states. This design not only streamlines the learning process but also fosters better generalization by reducing the tendency to overfit. Our works presents three key contributions: first, the development of the STN as a more effective alternative to MLPs; second, empirical evidence showing that STNs achieve superior sample efficiency in various training scenarios; and third, experimental results demonstrating that STNs match MLP performance in familiar environments and significantly outperform them in new settings. By embedding domain-specific knowledge, the STN enhances the Proximal Policy Optimization (PPO) algorithm's effectiveness without compromising performance, suggesting its suitability for a wide range of queueing network control problems.
\end{abstract}

\section{Introduction}


In the realm of queueing network controls, Deep Reinforcement Learning (DRL) has emerged a promising approach to learn high-performing network control policies \cite{dai_queueing_2022, ayesta_reinforcement_2022, raeis_queue-learning_2021, fawaz_deep_2021, wigmore_intervention-assisted_2024}.  As is standard in the DRL community, these works employ the Multi-layer Perceptron (MLP) neural network architecture for its flexibility and ability to serve as a universal function approximator \cite{pinkus_approximation_1999}.  However, this flexibility comes with significant drawbacks, particularly in terms of sample efficiency and generalization capabilities.  

Training MLP policy networks in DRL frameworks often requires a substantial number of interactions with the training environment(s) before learning a policy that performs well in the training environment.  Furthermore, the policies learned through this architecture tend to overfit to the training environments, resulting in poor performance when applied to new, unseen environments. This overfitting issue poses a major challenge, as real-world stochastic networks are dynamic and varied, necessitating robust and generalizable control policies. 

To address these challenges, we propose a novel policy network architecture tailored specifically for a large class of queueing network control problems: the switch-type neural network (STN).  Our approach is inspired by structural patterns observed in many non-learning policies within the network control literature, which often exhibit ``switch-type" behavior \cite{hsu_scheduling_2018}.  Specifically, these policies maintain consistency in their action selection across similar states, ensuring that if action $i$ is chosen for state $(s_1, ..., s_i, ..., s_K)$, then action $i$ will be chosen for state $(s_1, ..., s_i+1, ..., s_K)$, where the state element $s_i$ may represent the queue size at a node or channel conditions.  

By incorporating this switch-type structure into the neural network design, we aim to enhance both the sample efficiency during training and the generalization performance post training. Sample efficiency is improved because we restrict the search space to switch-type policies during training, rather than the class of all policies that can be represented by an MLP-based policy network.   Generalization performance is improved because we limit the ability of the neural network to memorize the training data, and instead learn a policy with consistent structure across the state-space. 

Our contributions are threefold: (1) we introduce a novel switch-type policy architecture for queueing network controls as an alternative to the MLP policy network architecture, (2) we provide extensive empirical validation demonstrating improved sample efficiency in both single-environment and multi-environment training scenarios, and (3) we conduct thorough experimental validation showing that trained switch-type policy networks match the performance of the MLP policy architecture when tested on the same environments seen during training, and significantly outperform the MLP architecture when tested on unseen environments.

\section{Background}
We start by providing an overview of the resource allocation problems of interest.  We then formally define switch-type policies and provide numerous examples in the literature of network control policies that are switch-type.  We then provide background on DRL and the MLP architecture. 

\subsection{Resource Allocation Problems}
We are going to focus on discrete time resource allocation where a network control agent chooses to interact with one of $K$ components (queues, servers, users, etc) in each time-step $t\in(0,1,2,...)$.  The state of component $k$,  $\mathbf s_{k,t}$, evolves according to one of two Markov transition distributions  $P(\mathbf S_{k,t+1}|\mathbf s_{k,t}, a_t=k)$ or $P(\mathbf S_{k,t+1}, \mathbf s_{k,t}, a_t\neq k)$ where $\mathbf S_{k,t+1}$ is a random variable representing the next state and $a_t$ is the ``action" or decision made by the central network controller in time $t$.  The objective is to minimize the average expected cost:
\begin{align}\label{eq: avg_cost}
    \lim_{T\rightarrow \infty}\frac{1}{T}\sum_{t=0}^{T-1}\mathbb E [c(\mathbf S_t)] =  \lim_{T\rightarrow \infty}\frac{1}{T}\sum_{t=0}^{T-1}\sum_{k=1}^{K}\mathbb E [c(\mathbf S_{k,t})]
\end{align}
where $\mathbf S_t = (\mathbf S_{1,t}, ..., \mathbf S_{K,t} )$ represents the network state at time $t$, and $c(\mathbf s_{k,t})$ is a cost function.  This problem can be modeled as a Markov Decision Process (MDP) $\mathcal M=(\mathcal S, \mathcal A, P, c, \rho_0)$ where the state-space $\mathcal S$ is the set of all possible networks states, $\mathcal A$ is a finite set of actions that can be taken in each state, $P(\mathbf S_{t+1}|\mathbf s_t,  a_t) = \prod_{k=1}^{K}P(\mathbf S_{k,t+1}|\mathbf s_{k,t}, a_t)$ is the Markov transition probability function, $c(\mathbf s)$ the cost function, and $\rho_0(\mathbf s)$ is the initial state-distribution.  The objective is to design or learn a Markov policy $\pi:\mathcal S\mapsto\Delta(\mathcal A)$ where $\Delta(\mathcal A)$ is a probability simplex over $\mathcal A$ as to minimize \cref{eq: avg_cost}.  For deterministic policies we use $\pi(\mathbf s)=i$ to denote that the policy $\pi$ chooses action $a=i$ in state $\mathbf s$, and for stochastic policies we use $\pi(i|\mathbf s)=p_i$ to denote that the policy $\pi$ chooses action $a=i$ with probability $p_i$ in state $\mathbf s$.  For a stochastic policy $\pi(\cdot|\mathbf s)$ denotes the distribution over all actions for policy $\pi$ in state $\mathbf s$. 

We provide two examples of such resource allocation problems below.
\subsubsection{Single-hop Scheduling}
\begin{figure}
    \centering
    \includegraphics[width=0.5\linewidth]{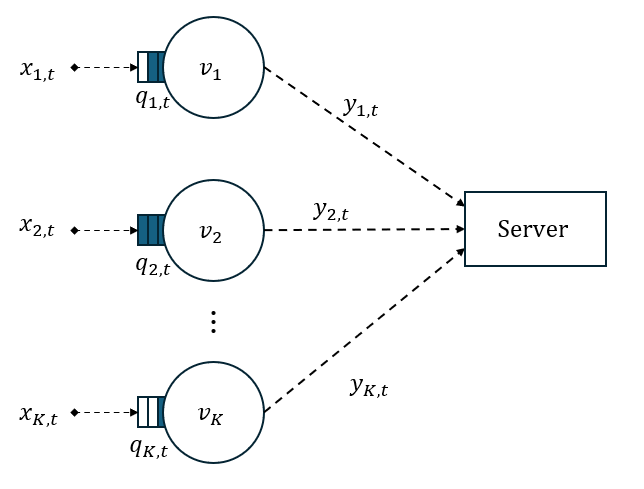}
    \caption{Example of a single-hop scheduling environment.  Packets arrive to each queue according to their independent arrival distributions.  The service rate between each queue and the server varies according to an independent i.i.d. process. 
    The server selects one of the $K$ queues to serve in each time-step. }
    \label{fig:sh_env}
\end{figure}

In the single-hop scheduling problem, there are \( K \) parallel queues and a single server. Each queue has a dedicated link to the server. Packets arrive at queue \( k \in [K] \) (where \([K] = \{1, 2, ..., K\}\)) at time \( t \) according to a discrete independent and identically distributed (i.i.d.) arrival distribution \( P(X_k) \), with \( X_k \) representing the number of arrivals for queue \( k \). Upon arrival at queue \( k \), packets are buffered, and the length of queue \( k \) at the beginning of time \( t \) is denoted as \( q_{k,t} \).

The capacity between queue \( k \) and the server also follows a discrete i.i.d. distribution \( P(Y_k) \), where \( Y_k \) represents the number of packets the server can process from queue \( k \). The arrival and capacity distributions are assumed to be independent of each other and the network states.The state of the queueing network at time \( t \) can be represented as a set of queue-length and capacity tuples \(\mathbf{s}_t = \{(q_{k,t}, y_{k,t})\}_{k \in [K]}\), with \(\mathbf{s}_{k,t} = (q_{k,t}, y_{k,t})\) denoting the state of queue \( k \) at time \( t \). At each time step, the agent observes the current network state \(\mathbf{s}_t\) and chooses a queue to serve, denoted by an integer \( a_t \in [K] \). The state of queue \( k \) evolves according to:

\[
q_{k,t+1} = 
\begin{cases} 
[q_{k,t} - y_{k,t}]^{+} + x_{k,t}, & \text{if } a_t = k \\ 
q_{k,t} + x_{k,t}, & \text{if } a_t \neq k 
\end{cases}
\]

where \([z]^{+} = \max(z, 0)\) and \((y_{k,t}, x_{k,t})\) are sampled i.i.d. from \((P(Y_k), P(X_k))\) respectively. This single-hop scheduling environment model is representative of networking scenarios where a shared resource must be allocated among \( K \) users, such as bandwidth in a wireless network or processor sharing.

\subsubsection{Multi-path Routing}
\begin{figure}
    \centering
    \includegraphics[width=0.6\linewidth]{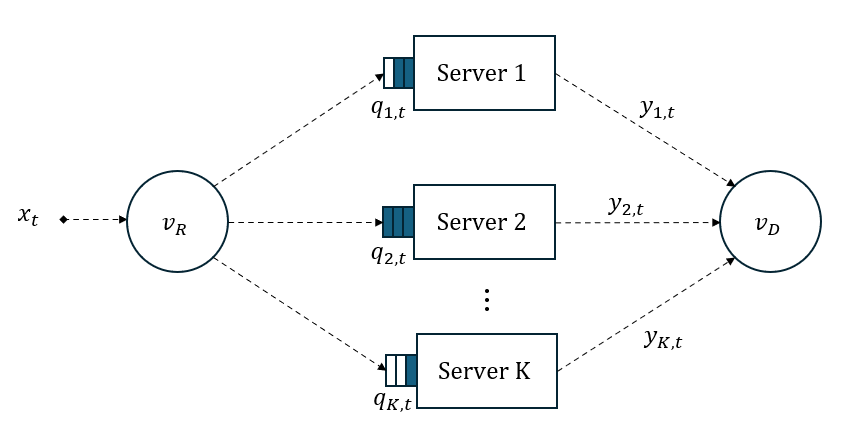}
    \caption{Example of a multi-path routing environment. A single packet enters the network at the routing node labeled $v_R$. The policy determines which of the $K$ queues the packet is then routed to. Each server maintains its own seperate queue.  After a packet is processed by any server, it is sent to node $v_D$ where it immediately leaves the network. }
    \label{fig:enter-label}
\end{figure}

In the multipath routing problem, there are \( K \) parallel servers, each with its own independent queue. Each server's service capacity follows a distribution \( P(Y_k) \), where \( y_{k,t} \sim P(Y_k) \) denotes the number of packets the \( k \)-th server can process from its queue at time \( t \). Packets exit the network immediately after being processed by their respective server. 

At each time-step, a single packet \( x_t \) arrives at a routing node. Similar to the single-hop scheduling environments, the network state is represented as \(\mathbf{s}_t = \{(q_{k,t}, y_{k,t})\}_{k \in [K]}\). At each time-step, a centralized controller selects a server to send the arriving packet to, denoted by \( a_t \in [K] \). The state of server \( k \)'s queue evolves according to:

\begin{align}
    q_{k,t+1} = 
\begin{cases} 
[q_{k,t} - y_{k,t}]^{+} + x_{t}, & \text{if } a_t = k \\ 
[q_{k,t} - y_{k,t}]^{+}, & \text{if } a_t \neq k 
\end{cases}
\end{align}
where \([z]^{+} = \max(z, 0)\).

\subsection{Switch-type Policy}
For resource allocation problems where the state $\mathbf s$ can be factored into $K$ components $\mathbf s = (\mathbf s_1, ... \mathbf s_K)$, we define the switch-type policy as follows:
\begin{defn}
    A switch-type policy is a deterministic Markov policy $\pi_{sw}:\mathcal S\mapsto \mathcal A$ that adheres to the following structure:
    If $\pi_{sw}(\mathbf s)=i$, for $\mathbf s = (\mathbf s_1, ...,\mathbf s_i, ... \mathbf s_K)$ where $\mathbf s_k\in \mathbb R^{N}$ then for any state $\mathbf s'$ satisfying:
    \begin{align}
        \mathbf s_k' = \begin{cases}
            \mathbf s_k + \mathbf b, & k=i \\
            \mathbf s_k, & k\neq i
        \end{cases}
    \end{align}
    for $\mathbf b\in (0,\infty)^{N}$, it follows that $\pi_{sw}(\mathbf s') = i$. 
\end{defn}
For instance, consider a single hop scheduling problem with $K=2$ queues. In a state $\mathbf s = (q_1,y_1,q_2,y_2)$, where $q_i$ denotes length of queue $i$ and $y_i$ denotes the capacity of the link between queue $i$ and the server, suppose that the switch type policy selects to serve queue $1$, indicated as $\pi_{sw}(\mathbf s)=1$. 
).  For the modified state $\mathbf s'=(q_1, y_1, q_2, y_2)$, where the length of queue $1$ has increased from $\mathbf s$ while all other elements remain the same, the switch type policy would also choose to serve queue $1$ i.e. $\pi_{sw}(\mathbf s')=1$.    Switch-type policies may also be switch-type in terms of some transformation of the state. For example, for a  multipath routing problems, we will focus on policies that are switch-type in terms of the network state given as $\mathbf s = (-q_1, y_1, -q_2, y_2)$. Meaning if the network control agent chooses to route the incoming packet to queue $1$ in state $\mathbf s = (-q_1, y_1, -q_2, y_2)$, it will also route the incoming packet to queue $1$ in the state $\mathbf s' = (-(q_1-1), y_1, -q_2, y_2)$  where the length of queue $1$ has been decreased.  When the decision regions are visualized, switch-type policies have clear ``switching curves" which indicate where the policy switches its decision from one action to another.  An example of these switching curves is shown in \Cref{fig: MDP}.

Switch-type policies are very prevalent in the network control literature. For example, MaxWeight style algorithms are very common for scheduling problems \cite{tassiulas_dynamic_1993, meyn_stability_2009, kadota_minimizing_2021, liu_tracking_2022, stolyar_maxweight_2004}. For these algorithms a weight $h(\mathbf s_k)$ is computed for each component state $\mathbf s_k\in \mathbf s$, and the MaxWeight policies chooses the action $\pi_{mw}(\mathbf s) = \argmax_k h(\mathbf s_k)$. When the weight function $h(\cdot)$ is non-decreasing in the component states $\mathbf s_k$, the MaxWeight policy is a switch-type policy.   For example, the function $h(\mathbf s_k)=q_k\times y_k$ is often used for the single-hop scheduling problem, meaning $\pi_{mw}$ is switch-type. The join-the-shortest-queue (JSQ) policy can similarly be thought of as a MaxWeight-type policy for a weighting function $h(\mathbf s_k)=-q_k$ for the multi-path routing problem \cite{lin_analysis_1992}. MaxWeight-style policies are often throughput optimal for their class of problems, meaning they can stabilize the queueing network if the network is stabilizeable \cite{neely_stochastic_2010}.  This makes them a particular useful class of policies as they have stability guarantees without requiring knowledge on the dynamics of the networks, however, they may be sub-optimal in terms minimizing costs.  

Whittle Index Policies are another common class of policies used in the resource allocation literature that are often switch-type \cite{whittle_restless_1988, nino-mora_dynamic_2007,okeeffe_whittles_2003, avrachenkov_congestion_2013} . Whittle Index policies are heuristic solutions for Restless Multi-Armed Bandit (RMAB) problems that satisfy a special property called indexability.  These policies compute a Whittle index $I(\mathbf s_k)$ for each arm's (component's) state, and select $\pi_{wi}(\mathbf s)=\argmax_k I(\mathbf s_k)$. When the component cost function $c(\mathbf s_k)$ is monotonically non-decreasing in $\mathbf s_k$, so is the Whittle Index $I(\mathbf s_k)$, meaning the resulting index policy is switch-type.  A major drawback of this approach is that indexability is often difficult to establish \cite{liu_indexability_2010}. Additionally, computing a closed form expression for the Whittle Index can be complex \cite{ayesta_unifying_2019}.  

Finally, there are cases in the literature where it can be shown that the optimal policy is switch-type.  For single-queue, multi-server scheduling problems it has been shown that the delay optimal policy is switch-type \cite{lin_optimal_1984, koole_simple_1995, viniotis_extension_1988}. This result was extended for multi-queue, multi-server MDPs in \cite{rykov_monotone_2001}.  For $N$-model networks, \cite{dai_queueing_2022} empirically demonstrates that the delay optimal policy is switch-type.  For a simplified version of the multi-user, single-server MDP where capacities are restricted to $y_k\in\{0,1\}$, and under the case that all capacity distributions are identical, \cite{tassiulas_dynamic_1993} proves that the Longest-Connected-Queue policy is delay optimal. Optimal switch-type policies are also not restricted to the objective of delay minimization.  In \cite{hsu_scheduling_2018}, the authors prove the optimal policy with regards to minimizing the sum of Age-of-Information (AoI) is also switch-type.  

\subsubsection{Empirical Demonstration of Switch-type Policy}
To illustrate the decision regions of a switch-type policy, we use the Policy Iteration algorithm to approximate an optimal policy for a specific single-hop scheduling problem instance \cite{bertsekas_dynamic_nodate}. First, we describe the single-hop scheduling problem of interest. This instance features two queues (\(K=2\)), each with Bernoulli arrivals at a rate parameter \(\lambda=0.4\). The service distributions for each queue is given by $P(Y_k=0)=0.5$, $P(Y_k=1)=0.3$, and $P(Y_{k}=2)=0.2$ for both queues.
This represents a symmetric problem where the arrival and service distributions are identical for each queue. Using these dynamics, we can create the transition function \(P(\mathbf{S}_{t+1}|\mathbf{s}_t, a_t)\) for any given state \(\mathbf{s}_t\) and action \(a_t\). Since the state-space is unbounded, we apply the sequence of approximate MDPs approach for MDPs with unbounded state-spaces \cite{sennott_average_1989}. The main idea is to solve for the optimal policy in sequentially larger finite-state approximations of the true MDP until these approximations converge over a predefined state-space region. We halt this process when the policy derived from policy iteration, \(\pi_{PI}\), stabilizes between iterations over the region \((q_1, q_2)\in[0, 20]\cap (y_1, y_2)\in[0,2]\). This approximates the true optimal policy over this truncated region of the state space.

\begin{figure}
    \centering
    \begin{subfigure}[b]{0.24\textwidth}
    \includegraphics[width = \textwidth]{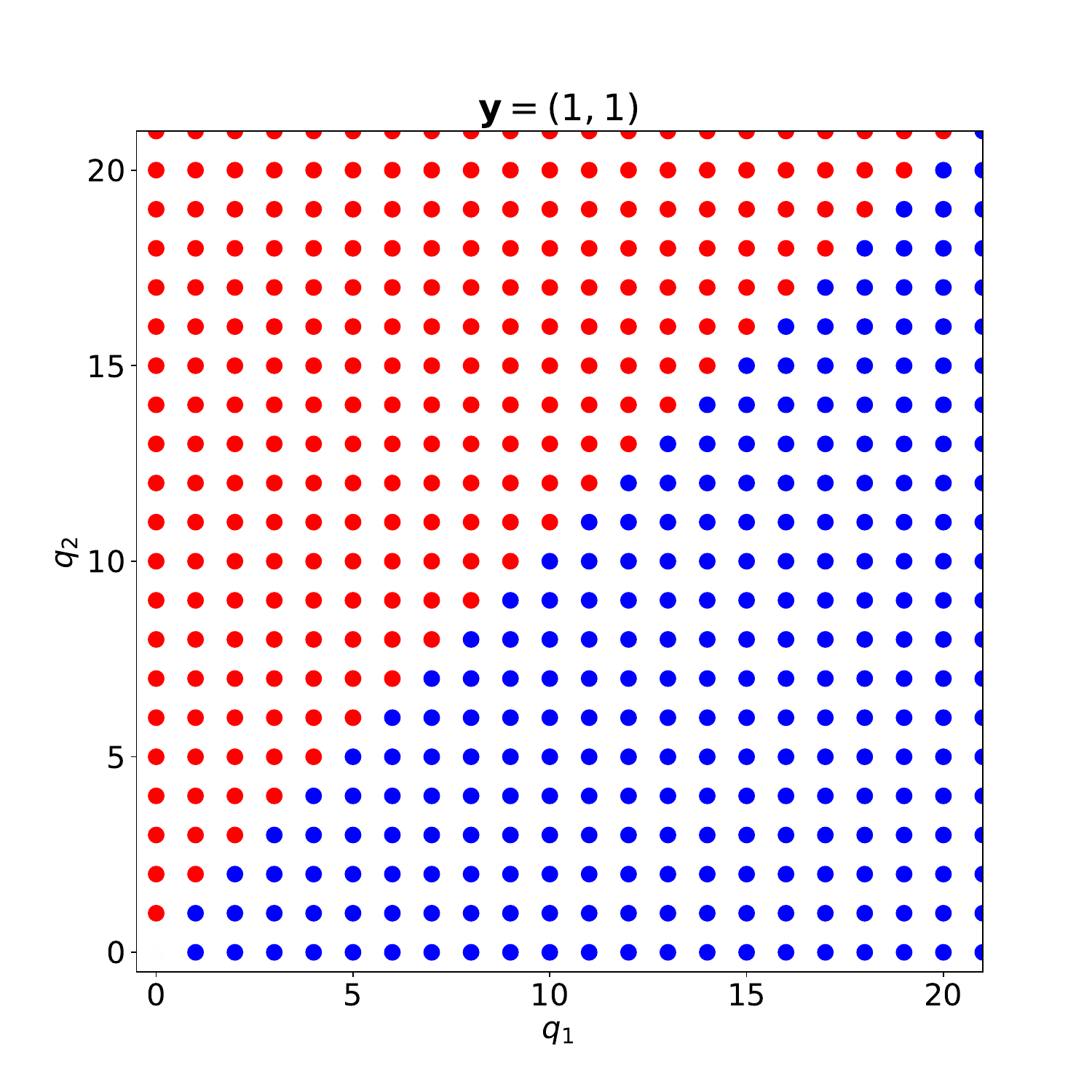}
    \caption{}
    \label{fig: MDP1}
    \end{subfigure}\hspace{-0.5em}
    \begin{subfigure}[b]{0.24\textwidth}
    \includegraphics[width = \textwidth]{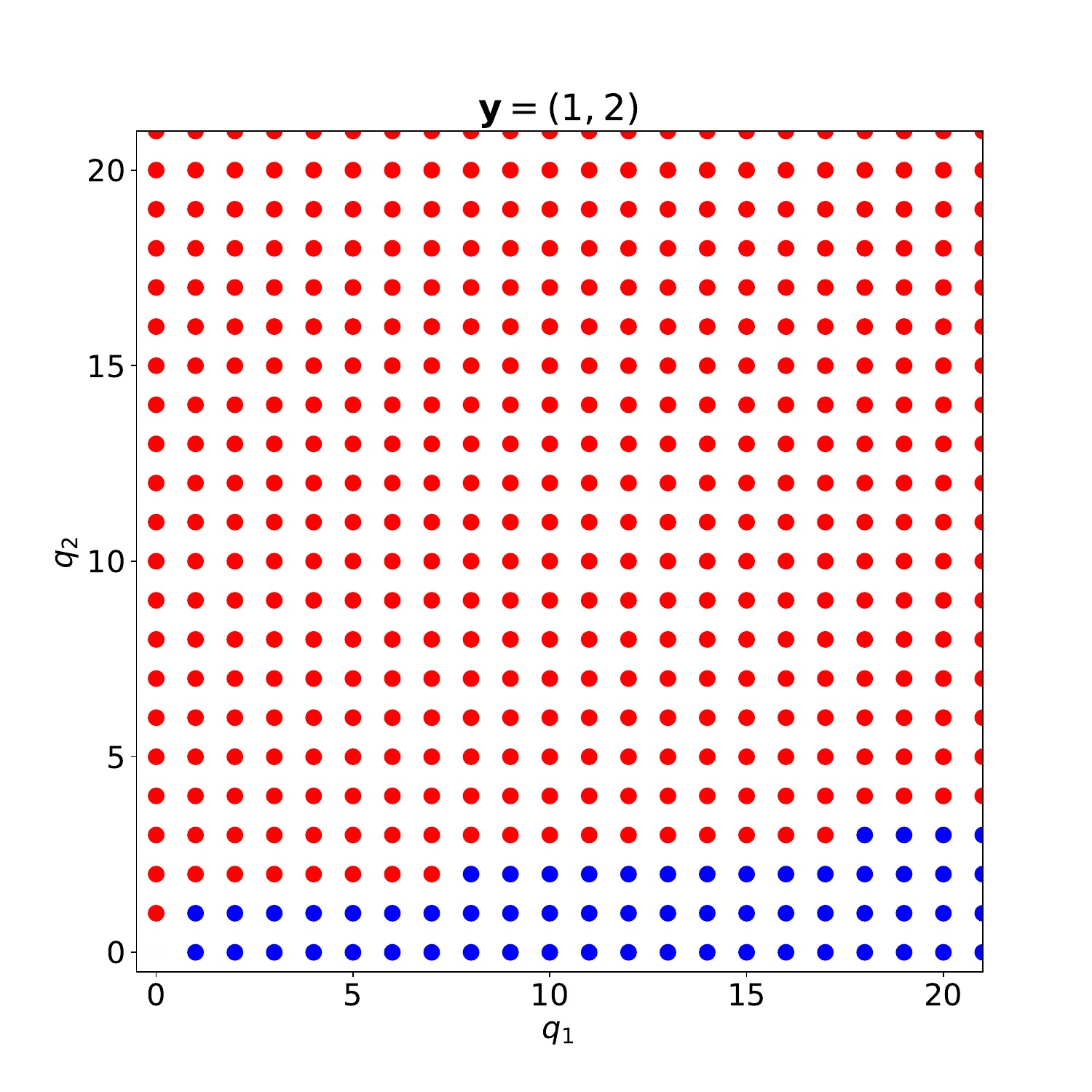}
    \caption{}
    \label{fig: MDP2}
    \end{subfigure}
    \begin{subfigure}[b]{0.24\textwidth}
    \includegraphics[width = \textwidth]{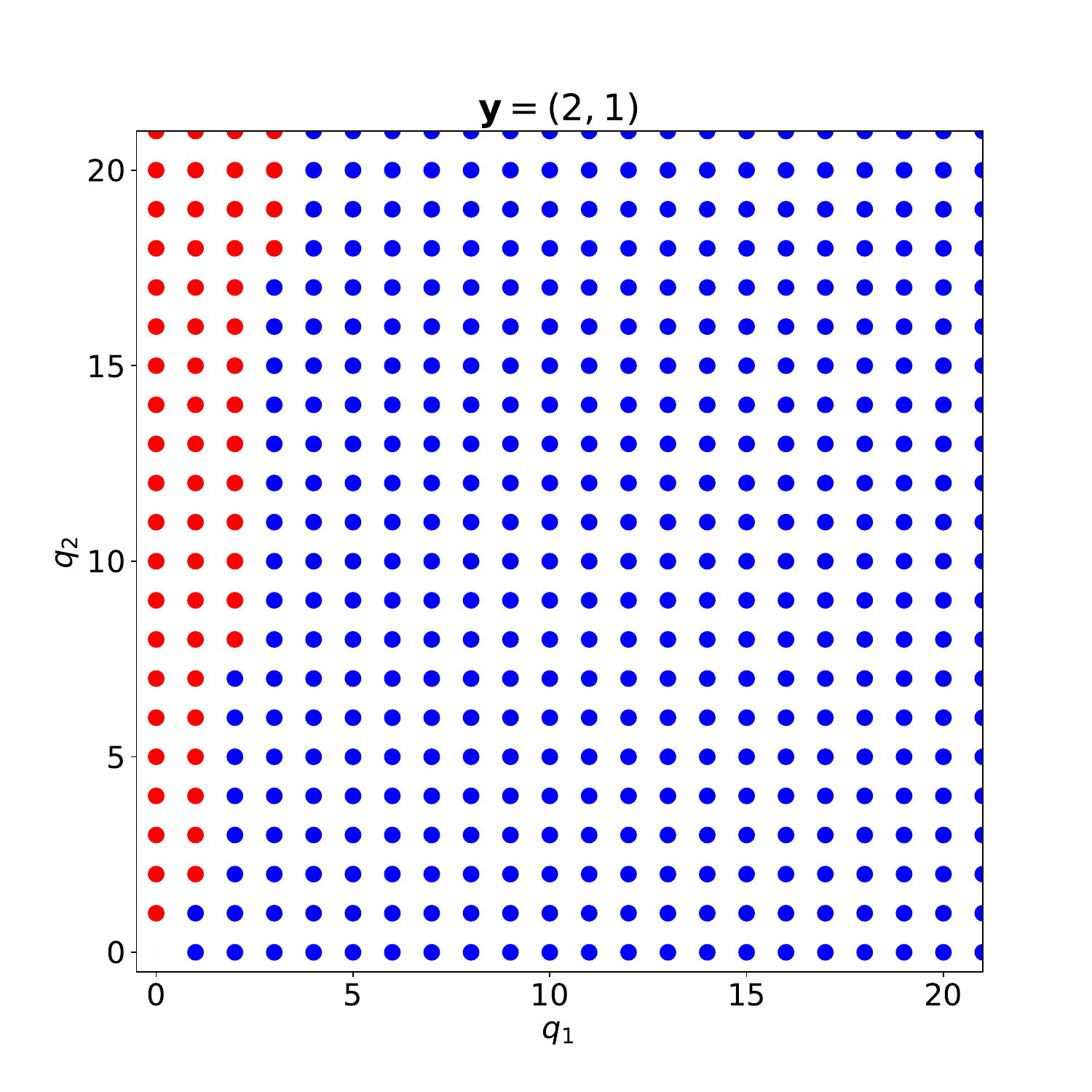}
    \caption{}
    \label{fig: MDP3}
    \end{subfigure}\hspace{-0.5em}
    \begin{subfigure}[b]{0.24\textwidth}
    \includegraphics[width = \textwidth]{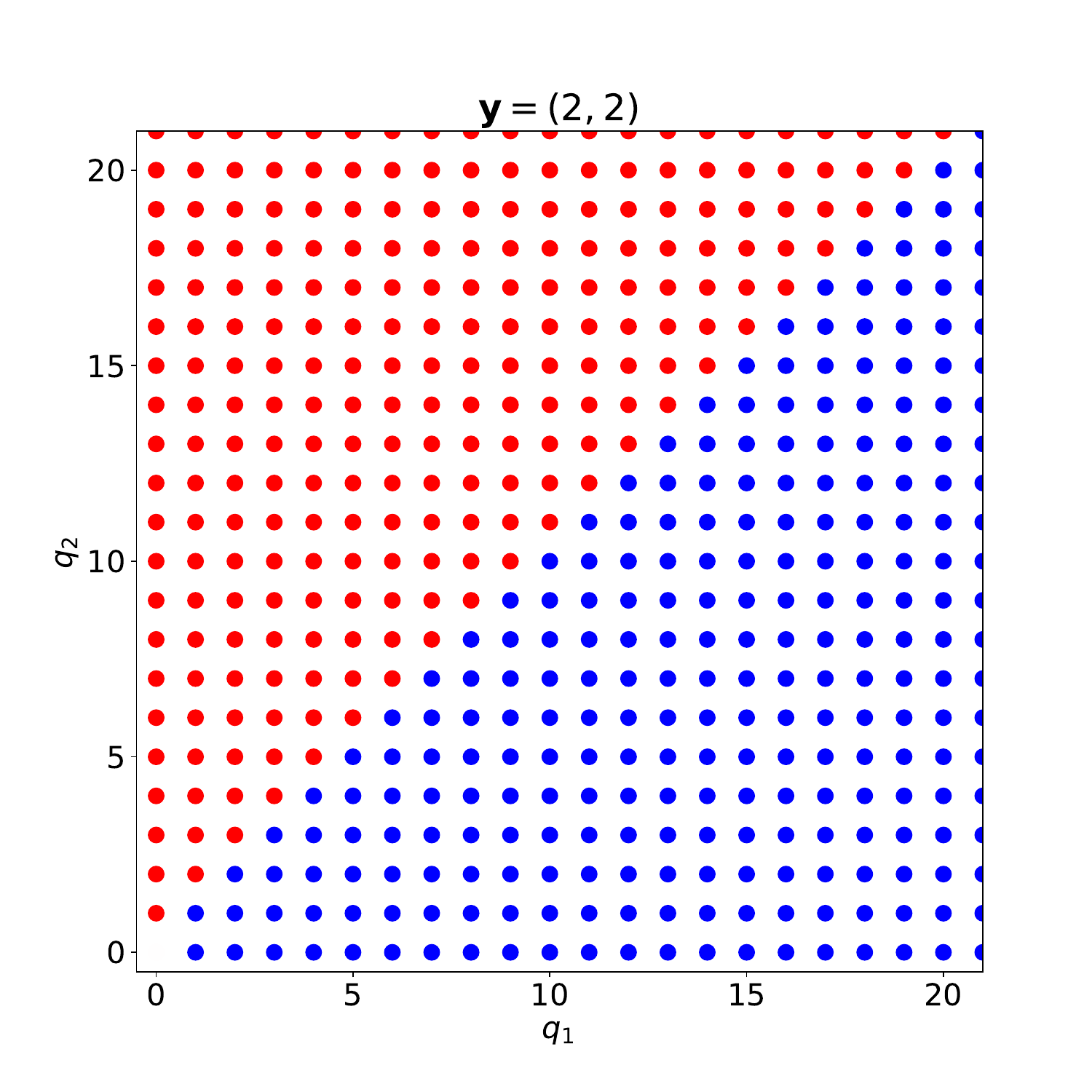}
    \caption{}
    \label{fig: MDP4}
    \end{subfigure}

    \caption{Decision regions for the policy $\pi_{PI}$.  Blue denotes $\pi_{PI}(\mathbf s)=1$ and red denotes $\pi_{PI}(\mathbf s)=2$.  The x-axis corresponds to $q_1$ and the y-axis corresponds to $q_2$. Each plot corresponds to a different set of service states $\mathbf y = (y_1, y_2)$. Thus the set of plots represents the decision region over the truncated state-space of $(q_1, q_2)\in[0, 20]\cap (y_1,y_2)\in[1,2]$}
    \label{fig: MDP}
\end{figure}

The decision regions of \(\pi_{PI}\) are visualized in \Cref{fig: MDP}. Figures \ref{fig: MDP1}-\ref{fig: MDP4} depict the decision regions over \((q_1, q_2)\in[0,20]\) with \((y_1, y_2)= (1,1), (1,2), (2,1)\), and \((2,2)\), respectively. In the symmetric network, when the link-states are equal, the optimal policy serves the longest queue, resulting in a linear switching curve. Conversely, in states where one service rate is higher than the other, the switching curve becomes nonlinear. Nonetheless, the policy consistently satisfies the conditions of a switch-type policy. This can be validated by selecting any point \((q_1, q_2)\) on any scatter plot. If the point is blue, meaning action \(\pi_{PI}(\mathbf{s})=1\) for the corresponding state, all points directly to the right will also be blue. Similarly, if the point is red, then all points directly above will be red. Switch-type in terms of $\mathbf y = (y_1, y_2)$ can be confirmed by picking an point $(q_1,q_2)$ and tracking how the corresponding changes between the different scatter plots representing the different realizations of $(y_1, y_2)$. For example, for any point in \Cref{fig: MDP1} such that $\pi_{PI}(q_1, q_2, y_1=1, y_2=1)=1$, we would also have $\pi_{PI}(q_1, q_2, y_1=2, y_2=1)=1$ as seen in \Cref{fig: MDP2}.

\subsection{Deep Reinforcement Learning}
In the previous section, we have highlighted several traditional approaches for designing queueing network control policies that happen to be switch-type.  MaxWeight policies, while effective in certain scenarios, often fail to minimize the average cost objective. Whittle index policies, although powerful, are restricted to indexable problems and deriving a closed-form expression for the Whittle index can be challenging. Dynamic programming and other model-based approaches are computationally feasible only for small-scale problems, making them impractical for larger, more complex networks. 

To address these limitations, we leverage model-free Deep Reinforcement Learning (DRL) \cite{sutton_reinforcement_2018}. Model-free DRL methods do not require an explicit model of the environment. Instead, they learn policies directly from interactions with the environment, making them suitable for large and complex problems where model-based methods are infeasible. More specifically, in this work we leverage the Proximal Policy Optimization (PPO) algorithm \cite{schulman_proximal_2017} for learning queueing network control policies which works as follows.

PPO is an actor-critic method, meaning it maintains two neural network approximators an actor or policy network $\pi_\theta$ and critic or value network $V_{\phi}$, there $\theta$ and $\phi$ represent the parameters of the neural networks.  PPO alternates between sampling data through interaction with the environment and optimizing a surrogate objective function using this data.  The key innovation of PPO is the used of a clipped-objective function to ensure the new policy does not deviate too far from the old policy, thereby stabilizing training. This is achieved by defining the probability ratio between the new and old policy $u_t(\theta) = \frac{\pi_{\theta}(a_t|\mathbf s_t)}{\pi_{old}(a_t|\mathbf s_t)}$. The objective function to be maximized is the clipped surrogate objective:
\begin{align*}
    \mathcal L^{\text{CLIP}}(\theta) = & \quad \underset{\rho_{\pi_\theta}, \pi_\theta}{\mathbb E}\Big[\min\big(u_t(\theta)\hat A^{\pi_{\theta}}(\mathbf s_t, a_t),\\ & \quad \text{clip}(u_t(\theta), 1-\epsilon, 1+\epsilon)\hat A^{\pi_{\theta}}(\mathbf s_t, a_t)\big)\Big ]
\end{align*}
where $\hat A^{\pi_{\theta}}(\mathbf s_t, a_t)$ is the advantage estimate, $\epsilon$ is a hyperparameter that controls the clipping range, and the expectation is with respect to the steady-state distribution $\rho_{\pi_\theta}$ induced by the policy $\theta$. The clipping operation $\text{clip}(u_t(\theta), 1-\epsilon, 1+\epsilon)$ ensures the policy update does not deviate significantly from the old policy.

\subsection{Multi-layer Perceptron Policy Networks}
Multi-layer Perceptrons (MLP) are the standard neural network architecture used for policy networks for environments with vector states and discrete actions. Mathematically, an MLP can be represented as a sequential stack of linear transformations with non-linear activation functions.  A single fully-connected layer of an MLP network can be written as
\begin{align}
    \mathbf z^{(l)}=\sigma(\mathbf W^{(l)} \mathbf z^{(l-1)} + \mathbf b^{(l)}) 
\end{align}
where $\mathbf W^{(l)}$ is the weight matrix of the $l$th layer, $\mathbf z^{(l-1)}$ is the output of the previous layer, $\mathbf b^{(l)}$ is the bias vector of the $l$th layer, and $\sigma(\cdot)$ is a non-linear activation function.  We use $\theta=\{(\mathbf W^{(l)}, \mathbf b^{(l)})\}_{l\in[L]}$ to refer to the sets of all parameters  associated with the neural network.  Due to their architecture and the universal approximation theorem, MLPs have the capability to approximate any continuous function to an arbitrary degree of accuracy, given sufficient neurons and layers \cite{pinkus_approximation_1999}. This makes them well-suited for representing complex, non-linear relationships between the environment state and the desired action or value. As written above, fully connected layers are continuous functions from $\mathbb R^{d_{l-1}}\mapsto\mathbb R^{d_{l}}$ where $d_l$ is the output dimension of the $l$th layer.  Let $\mathbf z\in \mathbb R^{K}$ denote the output of an MLP.  For DRL, we would like to convert $\mathbf z$ a probability distribution. For a discrete distribution over $K$ potential actions, a softmax layer is applied to the output of the MLP: 
\begin{align}
    \text{SoftMax}(\mathbf z)_i = \frac{\exp({z_i})}{\sum_{j\in[K]}\exp(z_j)}
\end{align}
which converts any vector $\mathbf z=(z_1, ..., z_K)\in \mathbb R^{K}$ into a normalized probability distribution over $K$ elements. 

\subsection{Zero-Shot Generalization}
Zero-shot generalization is the ability of a trained policy \(\pi_{\theta}\) to perform effectively in environments it has never encountered during training. The term ``zero-shot'' indicates that the test environments were never seen during training, unlike "few-shot" learning, which involves an additional fine-tuning step on the test environments after an extensive initial training period on the training environments. Zero-shot generalization is often challenging because standard DRL algorithms and architectures tend to overfit the policy to the training environments \cite{zhang_dissection_2018, kirk_survey_2023, cobbe_quantifying_2018}. Various strategies have been investigated to enhance the generalization capabilities of trained agents. For a detailed review, see the survey by Kirk et al. \cite{kirk_survey_2023}. Generally, these strategies can be classified into those that modify the training environment, adjust the loss functions during training, or alter the architecture of the policy and/or value networks. In this work, we focus on approaches that improve generalization by modifying the architectures of the policy and value networks.

For vision-based tasks, visual-attention mechanisms have gained popularity as they enable agents to identify crucial information from the input space \cite{tang_neuroevolution_2020, tang_sensory_2021, wang_unsupervised_2021}. These techniques are based on the hypothesis that only specific features of the input are relevant for optimal decision-making, and focusing on these features in unseen inputs enhances generalization. A similar attention mechanism was proposed in \cite{zambaldi_deep_2018, zambaldi_deep_2019}, which  is aimed at learning relationships between distinct entities within the visual input. This relational reinforcement learning (RRL) approach demonstrated improved generalization performance in video game and navigational tasks. For environments representable as shortest-path problems, the authors in \cite{vlastelica_neuro-algorithmic_2021} proposed a neuro-algorithmic policy architecture that incorporates a time-dependent shortest path solver within a deep neural network. This architecture's combinatorial generalization capabilities enable the learned policy to generalize effectively to new, unseen environments.

\section{Switch-Type Neural Network Design}
This section introduces our proposed neural network architecture to be used as policy networks in DRL applications.  This architecture, termed a ''switch-type policy network", ensures the resulting policy is a stochastic switch-type policy which is described below.

\begin{defn}
    A stochastic switch-type policy is a stochastic Markov policy $\pi_{ssw}:
\mathcal S\mapsto \Delta(\mathcal A)$ that adheres to the following structure:
    If $\pi_{ssw}(i|\mathbf s)=p_i$, for $\mathbf s = (\mathbf s_1, ...,\mathbf s_i, ... \mathbf s_K)$ where $\mathbf s_k\in \mathbb R^{N}$ then for any state $\mathbf s'$ satisfying:
    \begin{align}
        \mathbf s_k' = \begin{cases}
            \mathbf s_k + \mathbf b, & k=i \\
            \mathbf s_k, & k\neq i
        \end{cases}
    \end{align}
    for $\mathbf b\in (0,\infty)^{N}$, it follows that $\pi_{ssw}(i|\mathbf s') \geq p_i$. 
\end{defn}
In other words, if we increase any element of $\mathbf s_i$, the probability of choosing action $a=i$ increases under a stochastic switch-type policy. Stochastic policies are favored over deterministic policies for on-policy DRL algorithms, as stochastic policies ensure exploration of the state and action spaces which is crucial for learning an optimal policy.  

To represent stochastic switch-type policies using neural networks, we utilize stacked monotonic hidden layers which are described in the next subsection.  We formally define monotonicity with respect to multivariate functions first:

\begin{defn}
Let $ f:\mathbb R^{n}\mapsto \mathbb R$ be a continuous differentiable multivariate function. The function $f(\mathbf x)$ is said to be monotonically non-decreasing on $\mathbf x$, if $\forall i, x_i\leq x_i', \forall j\neq i, x_j= x_j'\Rightarrow f(\mathbf x) \leq f(\mathbf x')$ holds for any two points $\mathbf x, \mathbf x'\in \mathbb R^{n}$.  
\end{defn}

Our policy network architecture is based on the following lemma:

\begin{lem}\label{lem:mono} 
    For an MDP with state-space $\mathcal S = \mathcal S_1\times...\times \mathcal S_K$, let the function $\mathbf f:\mathcal S\mapsto \mathbb R^{K}$ be decomposable such that $\mathbf f(\mathbf s)=(f(\mathbf s_1), ..., f(\mathbf s_K))$ for some function $f$.  If for all $k\in[K]$,  $f:\mathcal S_k\mapsto \mathbb R$ is monotonically non-decreasing in all elements of $\mathcal S_k$, then the policy: 
    \begin{align} \label{eq:mono}
        \pi_\mathbf f(\cdot|\mathbf s) = \text{SoftMax} f(\mathbf s)
    \end{align}
    is a stochastic switch-type policy.  
    \end{lem}
Similarly, if softmax is replaced with an argmax, we obtain a deterministic switch-type policy.  

\subsection{Switch-Type Policy Architecture}
Our proposed architecture leverages two key components:
\begin{enumerate}
    \item \textbf{Monotonic Hidden Layers}: These hidden layers guarantee the resulting model is a monotonic function. This is achieved through a combination of exponentiated weights and a ReLU-$N$ activation function. 
    \item \textbf{Input Vector Transformation:} This component enables faster model evaluation during forward passes. 
\end{enumerate}
We delve into each component in the following subsections:

\subsubsection{Monotonic Hidden Layers}
Our switch-type policy network is comprised of fully-connected hidden layers that enforce monotonicity.  These monotonic hidden layers utilize an exponentiated weight unit and Relu-N activation function and this combination was first proposed for universal monotonic function approximation in \cite{kim_scalable_2023}.  Exponentiated weight units perform the following linear operation $$\exp(\mathbf W^{(l)}) \mathbf z^{(l-1)} + \mathbf b^{(l)}$$ where $\mathbf W^{(l)}$ and $\mathbf b^{(l)}$ is the weight matrix and bias vector of the $l$th layer, and $\mathbf z^{(l-1)}$ is the output of the previous layer. Since $\exp(\mathbf W^{(l)})$ is guaranteed to be positive, this linear operation is monotonically non-decreasing in $\mathbf z^{(l)}$. The output of this linear operation is then passed through a ReLU-N  activation function $\sigma_{RN}:\mathbb R^{d}\mapsto[0, N]^{d}$,  \cite{liew_bounded_2016} defined as:
\begin{align}
    \sigma_{RN}(\mathbf z) = \begin{cases}
        N, & \text{if } z_i >N, z_i\rightarrow N^+ \\
        x, & \text{if } 0 < z_i< N, z_i\rightarrow N^-, z_i\rightarrow 0^+\\
        0, & \text{if }z_i <0, z_i\rightarrow 0^-
    \end{cases}
\end{align}

This combination guarantees the output of the $l$th layer: 
\begin{align}
    \mathbf z^{(l)}=\sigma_{RN}(\exp(\mathbf W^{(l)})) \mathbf z^{(l-1)} + \mathbf b^{(l)}) 
\end{align}
is monotonically non-decreasing with respect to the previous layer's output $\mathbf z^{(l-1)}$.  The ReLU-N activation function is used instead of the commonly used ReLU activation function, as a ReLU network with all positive weights can only model a convex function \cite{liu_certified_2022-1}. Meanwhile, a fully-connected ReLU-N network with at least four hidden layers is a universal function approximator for purely monotonic functions \cite{kim_scalable_2023}. Much like an MLP architecture, these monotonic hidden layers can be stacked to create a monotonic deep neural network.  For an $L$ layer network, the final output layer is comprised of an exponentiated weight unit with a weight vector $\mathbf w_L$ and scalar bias $b_L$ without a final activation function, e.g., 
\begin{align}
 \mathbf  z^{(L)}= \exp(\mathbf W^{(L)})\mathbf z^{(L-1)}+ \mathbf b^{(L)}
\end{align}
This ensures the model formed by the stacked layers $$f_{\theta}(\mathbf z^{(0)})=\exp(\mathbf W^{(L)}) \sigma_{RN}(\exp(\mathbf W^{(l-1)})^\intercal \cdots )) + b^{(L)}))$$
where $\mathbf z^{(0)}\in \mathbb R^{n}$ is the input vector to the model, is a multivariate monotonic function $ f_\theta:\mathbb R^{n}\mapsto \mathbb R$. 

\subsubsection{Input Transformation}
The monotonic neural network $f_\theta$ takes the place of the monotonic function $f$ in  Lemma \ref{lem:mono}.  Unlike standard policy networks, $f_{\theta}$ is not applied to the entire state vector $\mathbf s$, but is instead applied to each component state $\mathbf s_k\in\mathbf s$ individually. This operation can be performed in parallel by a simple reshaping operation that takes observation $\mathbf s \in \mathbb R^{K\times n}$ and converts it to a matrix $\mathbf S\in \mathbb R^{(K,n)}$ where the $k$th row is $\mathbf s_{k}$.  Passing $\mathbf S$ through the monotonic neural network then produces a vector $\mathbf z^{L}\in \mathbb R^{K}$ where each component $z_k^{L}$ is monotonically non-decreasing in $\mathbf s_k$. 

\subsubsection{Output Layer}:
We add a softmax layer on top of the final monotonic hidden layer to produce the switch-type policy network. This resulting policy architecture ensures that for any weights $\theta=\{(\mathbf W^{(l)}, \mathbf b^{(l)})\}_{l\in[L]}$, the resulting policy $\pi_{\theta}$ is a stochastic switch type policy.

\section{Single Environment Training Comparison}
First we compare STN policy networks to MLP policy networks in terms of their sample efficiency during training. We focus on the case where a single policy is trained on a single environment.  These environments model single-hop scheduling or multi-path routing problems. 

\subsection{Training Environments}
First we describe the environment sampling procedure for the single-hop scheduling and multi-path routing environments.

\subsubsection{Single-hop Environment Sampling Procedure} \label{sec:sh_samp}
The following sampling procedure randomly generates a set of single-hop environments  \(\mathbfcal E^{(SH)} = (\mathcal E^{(1)}, \mathcal{E}^{(2)}, \ldots)\) where each $\mathcal E^{(j)}\in \mathbfcal E$ has a different mean arrival rate vector $\boldsymbol{\lambda}=(\lambda_k)_{k\in[K]}$ where $\lambda_k=\mathbb E[X_k]$ and a different service rate vector $\boldsymbol{\mu}=(\mu_k)_{k\in [K]}$.  For the experiments in this section, all single-hop environments \(\mathcal E^{(j)} \in \mathbfcal E^{(SH)}\) have four queues ($K=4$). The arrival distributions are Poisson with arrival rate parameter \(\lambda_k^{(j)} = \mathbb{E}[X_k]\) for queue $k$ of environment \(\mathcal E^{(j)}\). Similarly, the service distributions are Poisson with service rate parameter \(\mu_k^{(j)} = \mathbb{E}[Y_k]\) for queue $k$ of environment \(\mathcal E^{(j)}\). Each environment \(\mathcal E^{(j)}\) in \(\mathbfcal E^{(SH)}\) corresponds to a different set of arrival rate parameters \(\boldsymbol{\lambda}^{(j)} = \{\lambda_k^{(j)}\}_{k \in [K]}\) and service rate parameters \(\boldsymbol{\mu}^{(j)} = \{\mu_k^{(j)}\}_{k \in [K]}\). These environment parameters \((\boldsymbol{\lambda}^{(j)}, \boldsymbol{\mu}^{(j)})\) were generated as follows:

For each environment \(\mathcal E^{(j)}\), all parameters were sampled i.i.d. from a uniform distribution over the range (0,3). To ensure that \(\mathbfcal E^{(SH)}\) did not contain non-stabilizable environments, the MaxWeight policy \(\pi_0\) was run on \(\mathcal E^{(j)}\) for three separate trajectories of length \(T=50,000\) (\(\tau_1, \tau_2, \tau_3\)), each using a different random seed. For each trajectory, the empirical time-average cost was computed as $
\hat{J}(\pi_0, \tau) = \frac{1}{T} \sum_{\mathbf{s}_t \in \tau} c(\mathbf{s}_t)
$. 
These time-average costs were then averaged to produce \(J(\pi_0, \mathcal E^{(j)})\), which is an estimate of the true average cost. If \(J(\pi_0, \mathcal E^{(j)})\) was greater than 200, the environment parameters were resampled, as it was assumed that \(\mathcal E^{(j)}\) was not stabilizable. This process was repeated until five different stabilizable environments were obtained.

\subsubsection{Multi-path Environment Sampling Procedure} \label{sec:mp_samp}
The Multi-path routing environment sampling procedure follows the same steps as the single-hop sampling procedure, except only the service rates $\boldsymbol{\mu}^{(j)}$ are sampled for each environment as the arrival process is deterministic for this class of problems.  All environments generated have eight queues/servers ($K=8$) and each service distribution was Poisson with rate parameter $\mu_k$ sampled from a uniform distribution over the range (0,1). To determine if the parameters correspond with a stabilizable network, we use the Shortest Queue routing policy as $\pi_0$ and calculate $J(\pi_0, \mathcal E)$, and add $\mathcal E$ to the training set $\mathbfcal E^{(MP)}$if this empirical average cost is less than 200.  

\subsection{Training and Evaluation Procedure}
The following process was repeated for both sets of environments $\mathbfcal E^{(SP)}$ and $\mathbfcal E^{(MP)}$, and we will use $\mathbfcal E$ in this section to refer to either one of these sets.  This set $\mathbfcal E$ was generated using corresponding sampling procedure above, and it contained five different environments.  For this study, an STN policy network $\pi_{\theta, STN}^{(j)}$  and MLP policy network $\pi_{\theta, MLP}^{(j)}$ was trained on each environment $\mathcal E^{(j)}\in \mathbfcal E$ using the PPO algorithm.     
During training, we utilize the same hyperparameters for both network architectures, except for the learning rate. The optimal learning rates for each architecture are determined by sweeping over the range \(10^{-2}\) to \(10^{-5}\), and selecting the rate that results in the lowest average cost \({J}(\pi^{(j)}_{\theta}, \mathcal{E}^{(j)})\) during the evaluation phase over all training environments. These hyperparameters are given \Cref{tab:hyperparameters}, and all results correspond to the runs utilizing this learning rate.

\begin{table}
    \centering
    \begin{tabular}{|c|c|} \hline 
         \textbf{Hyperparameter}& \textbf{Value}\\ \hline 
         Learning Rate (STN)& 3e-3 \\ \hline 
 Learning Rate (MLP)&3e-4\\ \hline 
         Batch Size& 2,000\\ \hline 
 Mini-batch Size&100\\ \hline 
         Update Epochs/Batch& 3\\ \hline 
         $\lambda_{GAE}$& 0.95\\ \hline 
         $\epsilon$& 0.1\\ \hline 
         Entropy Coef.& 0.01\\ \hline
    \end{tabular}
    \caption{Hyperparameters for training both the STN and MLP policy networks. }
    \label{tab:hyperparameters}
\end{table}

Throughout the training process, we monitor the moving average cost over the past 5,000 time-steps as a running average estimate of \(J(\pi_{\theta}^{(j)}, \mathcal{E}^{(j)})\). After \(T_{\text{train}}=1,000,000\) training steps, we cease training and evaluate the trained policy on the same environment it was trained on. For evaluation, the trained policy \(\pi_{\theta}\) generates three 50,000-step trajectories in the same environment, each using a different random seed. Both the STN and MLP policy networks use the same three seeds for consistency.  From this evaluation phase we obtain the estimated time-average cost $J(\pi_\theta^{(j)}, \mathcal E^{(j)})$ for each training environment $\mathcal E^{(j)}\in \mathbfcal E$ and both policies $\pi_{\theta, STN}^{(j)}$ and $\pi_{\theta, MLP}^{(j)}$ that were trained on $\mathcal E^{(j)}$

During training, the neural network outputs are passed through a SoftMax layer to facilitate exploration. In the evaluation phase, however, an argmax layer is used instead, making each policy deterministic. This approach ensures that the evaluation accurately reflects the performance of the policy without the stochastic variations introduced during training.

\subsection{Single Environment Training Results}
Here we present the results from the single environment training experiments. The optimal average cost $\min_\pi J(\pi,\mathcal E^{(j)} )$ differs for each environment.  Thus for a better comparison of the performance we compute the normalized average cost defined as: 
\begin{align}
    J_0(\pi_{\theta}, \mathcal E^{(j)}) = \frac{J(\pi_{\theta},\mathcal E^{(j)})}{J(\pi_{0},\mathcal E^{(j)})}
\end{align}
where $\pi_0$ is the MaxWeight policy for single-hop scheduling or the Shortest Queue policy for the mult-path routing environments. The average cost $J(\pi_0, \mathcal E^{(j)} )$ is obtained from the environment sampling procedure.  

\begin{figure}
    \centering
    \begin{subfigure}[b]{0.24\textwidth}
    \includegraphics[width = \textwidth]{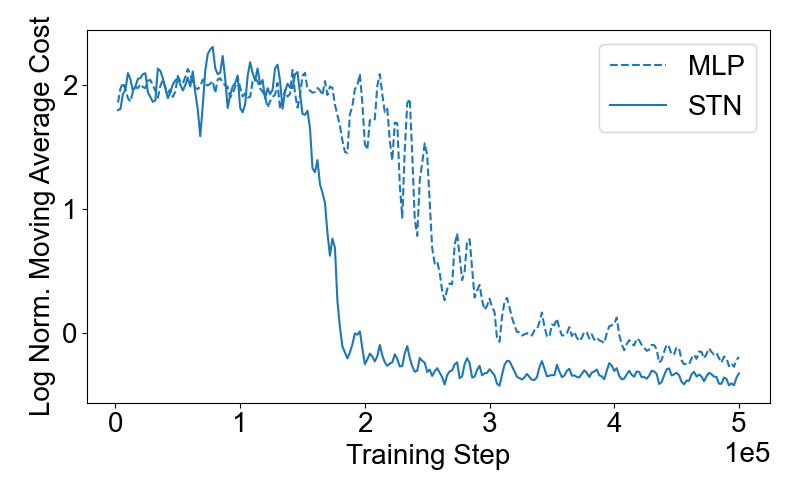}
    \caption{Environment SH1}
    \label{fig: SH_tr_1}
    \end{subfigure}\hspace{-0.4em}
    \begin{subfigure}[b]{0.24\textwidth}
    \includegraphics[width = \textwidth]{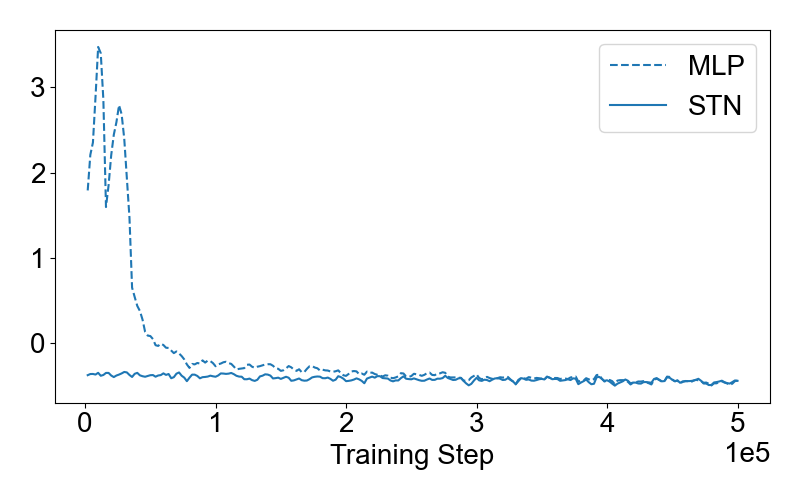}
    \caption{Environment MP2}
    \label{fig: MP_tr_1}
    \end{subfigure}\hspace{-0.4em}
    \begin{subfigure}[b]{0.24\textwidth}
    \includegraphics[width = \textwidth]{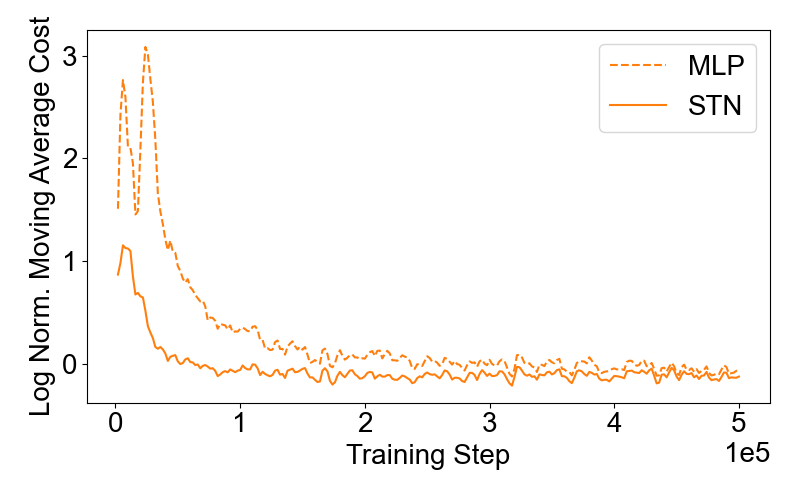}
    \caption{Environment SH2}
    \label{fig: SH_tr_2}
    \end{subfigure}\hspace{-0.4em}
    \begin{subfigure}[b]{0.24\textwidth}
    \includegraphics[width = \textwidth]{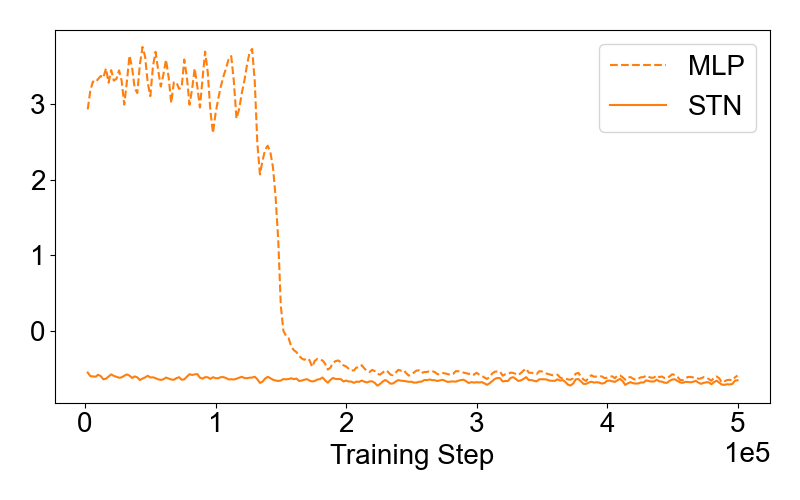}
    \caption{Environment MP2}
    \label{fig: MP_tr_2}
    \end{subfigure}\hspace{-0.4em}
    \begin{subfigure}[b]{0.24\textwidth}
    \includegraphics[width = \textwidth]{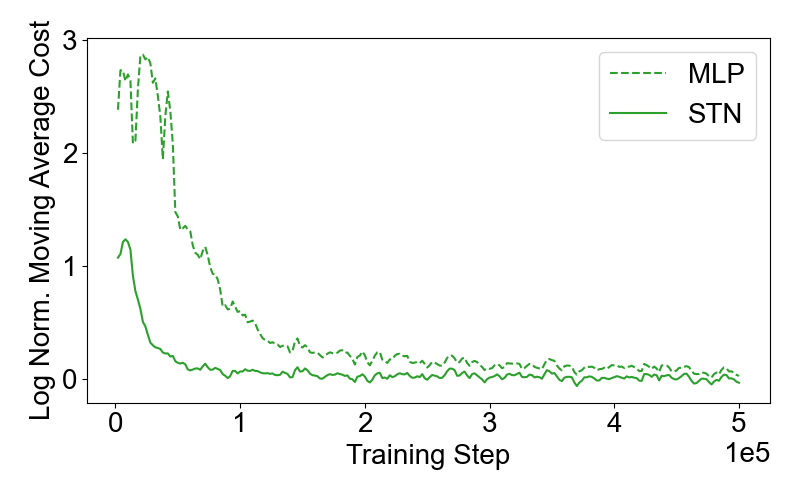}
    \caption{Environment SH3}
    \label{fig: SH_tr_3}
    \end{subfigure}\hspace{-0.4em}
    \begin{subfigure}[b]{0.24\textwidth}
    \includegraphics[width = \textwidth]{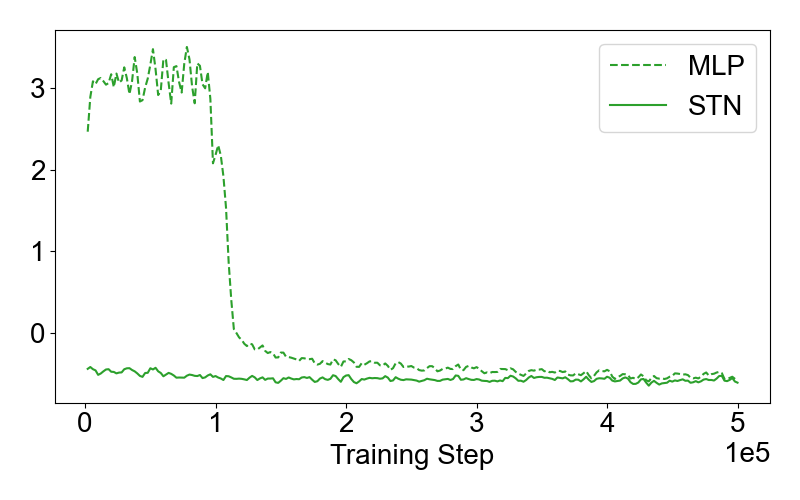}
    \caption{Environment MP3}
    \label{fig: MP_tr_3}
    \end{subfigure}\hspace{-0.4em}

    \begin{subfigure}[b]{0.24\textwidth}
    \includegraphics[width = \textwidth]{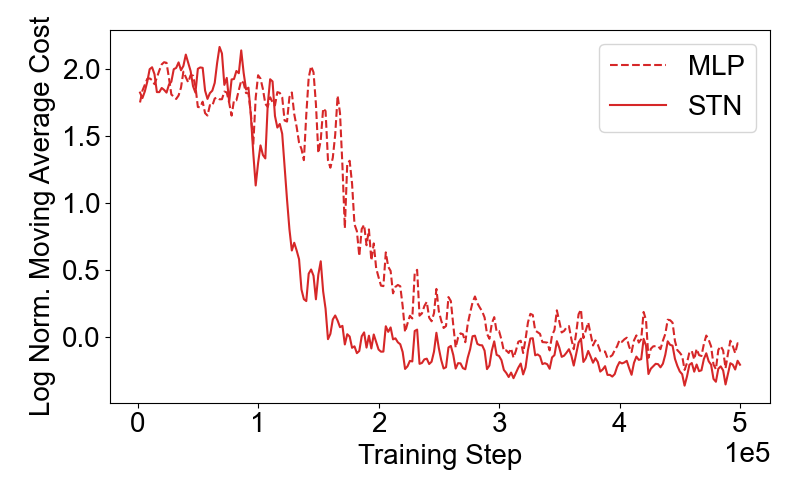}
    \caption{Environment SH4}
    \label{fig: SH_tr_4}
    \end{subfigure}\hspace{-0.4em}
    \begin{subfigure}[b]{0.24\textwidth}
    \includegraphics[width = \textwidth]{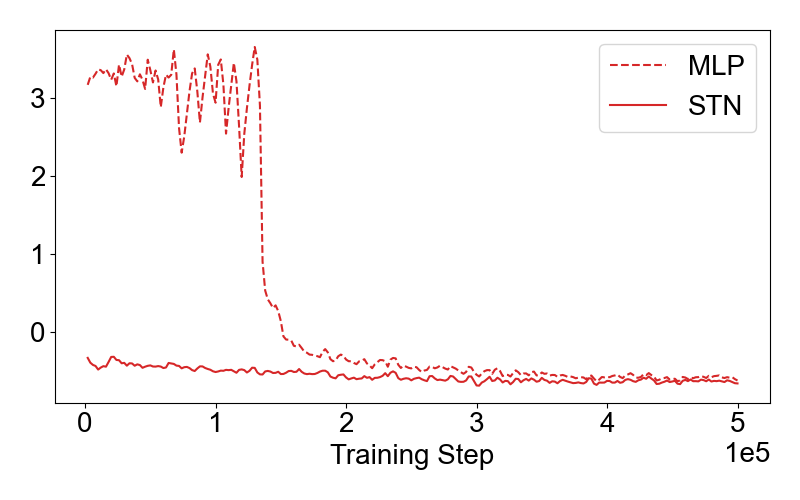}
    \caption{Environment MP4}
    \label{fig: MP_tr_4}
    \end{subfigure}\hspace{-0.4em}
    \begin{subfigure}[b]{0.24\textwidth}
    \includegraphics[width = \textwidth]{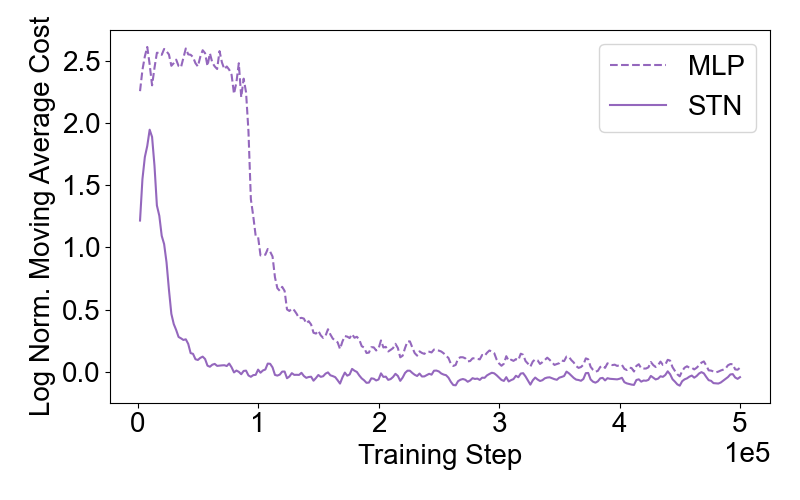}
    \caption{Environment SH5}
    \label{fig: SH_tr_5}
    \end{subfigure}\hspace{-0.4em}
    \begin{subfigure}[b]{0.24\textwidth}
    \includegraphics[width = \textwidth]{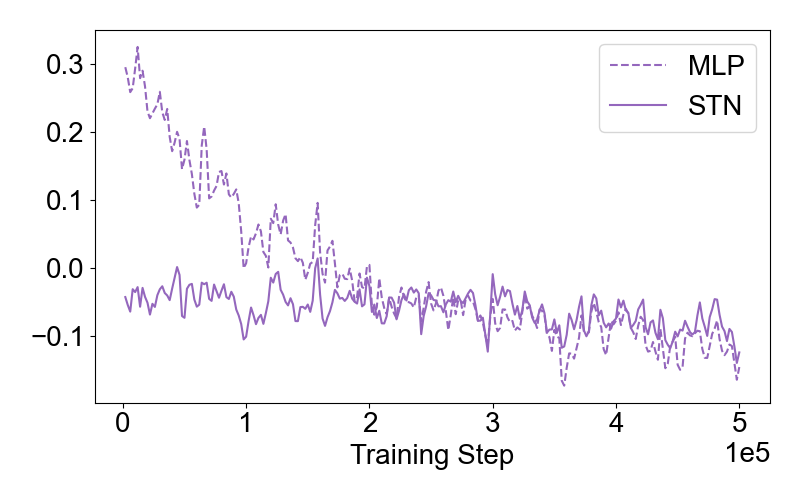}
    \caption{Environment MP5}
    \label{fig: MP_tr_5}
    \end{subfigure}\hspace{-0.4em}

    \caption{Moving average cost of the training policy versus training step.  The left column corresponds to the single-hop scheduling training environments and the right column corresponds to the multi-path routing training environments. }
    \label{fig:single_tr_curves}
\end{figure}

To compare the sample efficiency during training, we record a $T_{MA}=5,000$ step moving average of the cost during the training process for each policy/environment. Since the average cost is reduced significantly over the training period, we plot the natural logarithm of this moving average in \Cref{fig:single_tr_curves} for both problem classes. Since we are taking the natural log, the time-step in which log normalized moving average cost drops below 0.0 corresponds with the time during training in which the trained policy would outperform $\pi_0$.  Additionally, although training was done over a total of $T_{train}=1,000,000$ steps, we only plot the first $500,000$ steps as the policy performance converges before then and does not improve significantly after convergence. These results demonstrate that an STN policy network requires fewer samples from the environment to produce a policy that outperforms $\pi_0$ compared to the MLP policy architecture. Additionally, even before training, the STN policy architecture often results in a policy that outperforms the initial policy of the MLP policy network. This difference in initial policy performance is very apparent for multi-path routing environments as seen by the plots in the right column of \Cref{fig:single_tr_curves}.  In fact, the initial STN policy only slightly improves during training over training for the multi-path environments.

\begin{figure}
    \centering
    \begin{subfigure}[b]{0.24\textwidth}
        \includegraphics[width= \textwidth]{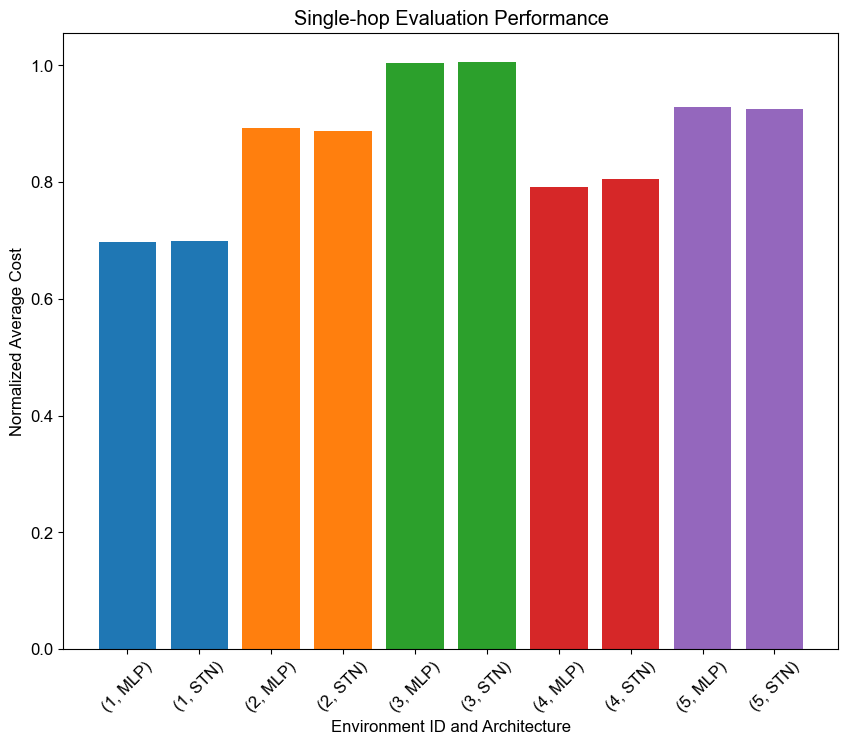}
    \end{subfigure}\hspace{-0.4em}
    \begin{subfigure}[b]{0.24\textwidth}
        \includegraphics[width=\textwidth]{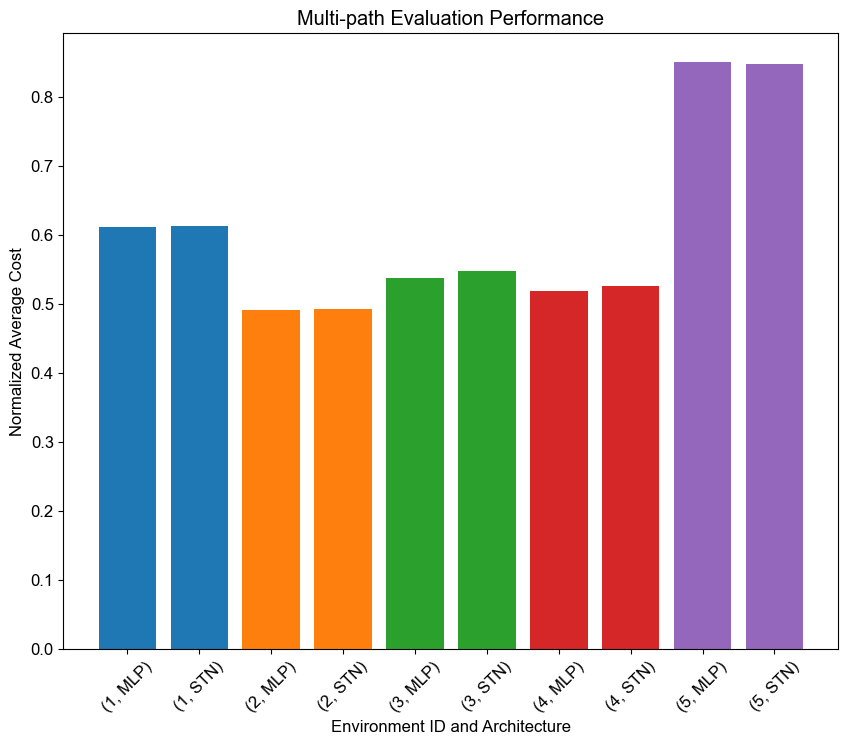}
    \end{subfigure}
    
    \caption{Average cost of the trained policies in the same environments they were trained in.  }
    \label{fig:Single_eval}
\end{figure}

In addition to measuring performance during training, we plot average cost of the trained policies on their respective environments in \Cref{fig:Single_eval}. From these plots, its apparent that the architectures produce policies with identical performance. This means that even though the MLP policy network architecture is optimizing over a very large class of policies, it converges to the same performance as optimizing over the class of switch-type policies.  We believe this indicates that the optimal policy is switch-type for these two resource allocation problems.

\section{Zero-Shot Generalization}
Next, we study the zero-shot generalization performance of the STN and MLP policy networks. The following experimental procedure was repeated for both the single-hop scheduling and multipath routing problem classes.   First, an environment set $\mathbfcal E$ containing 100 different environments was generated using the parameter sampling procedure detailed in Sections \ref{sec:sh_samp} and \ref{sec:mp_samp}.  This environment set $\mathbfcal E$ was then split into a training set $\mathbfcal E_{train}$ which contained five environments, and $\mathbfcal E_{test}$ which contained the remaining ninety-five environments from $\mathbfcal E$.  A single policy network was trained for each policy network architecture on all environments in $\mathbfcal E_{train}$, and this single policy network was then tested on all environments $\mathbfcal E$. The performance at test time on environments in $\mathbfcal E_{train}$ characterize how well the policy network learned to perform in environments it saw during training, while the performance on environments in $\mathbfcal E_{test}$ characterize a policy networks ability to generalize to unseen environments.  

\subsection{Training and Evaluation Procedure}
In the previous single-environment training procedure, we trained a single policy $\pi_{\theta}^{(j)}$ on each environment $\mathcal E^{(j)}$ in the training set. For the current study, we trained a single policy $\pi_{\theta}$ on all environments in the training set \(\mathbfcal E\) simultaneously. The training process is only slightly modified from the single-environment procedure: trajectories of length \(T_{eps} = 2000\) are generated using \(\pi_{\theta}\) in each environment in parallel, resulting in an update step with a batch size of 10,000 for each update. The total number of training steps collected during the training phase is 10,000,000, meaning that 2,000,000 training steps are collected from each training environment over the entire training process.

After training, we evaluated the trained policy on all environments in $\mathbfcal E$.  This was done by generating three trajectories using the trained policy on each environment using a different random seed for each trajectory.  

\subsection{Observation Encoding}
In the single-environment study, the policy was not required to observe the environment parameters in order to learn a policy that performs well.  However, for this study of generalization performance, since the objective is to perform well over the entire environment set $\mathbfcal E$, the agent must observe both the current network state $\mathbf s_t$, but also the current environments parameters.  Otherwise, the environment would only be partially observable, as there would be no way of the agent of knowing which environment its currently operating in. Thus, for this study we assume the agent observes not only the $(q_{k,t}, y_{k,t})$ for each queue $k$, but also the corresponding environment parameters i.e. $\lambda_k$ and $\mu_k$ for the single-hop scheduling environments and $\mu_k$ for the multipath routing problem.  However, due to the monotonic nature of the the STN policy network, the sign of the individual elements in the observation vector is important.  Thus in order to determine the appropriate sign, we trained an STN policy network on five different environments using different encoding of the environment parameters for both problem classes.  It was found that the observation encoding of $\mathbf o_{k,t}=(q_{k,t}, y_{k,t}, \lambda_k, -\mu_k)$ produced the best policy for single-hop environments and $\mathbf o_{k,t} = (-q_k, y_k, \mu_k)$ produced the best policy for the multi-path environments and thus both the policy and value networks utilized in this study used observation vectors encoded in this form. 

\subsection{Zero-Shot Generalization Performance}
\begin{figure}
    \centering
    \begin{subfigure}{0.4\textwidth}
        \includegraphics[width = \textwidth]{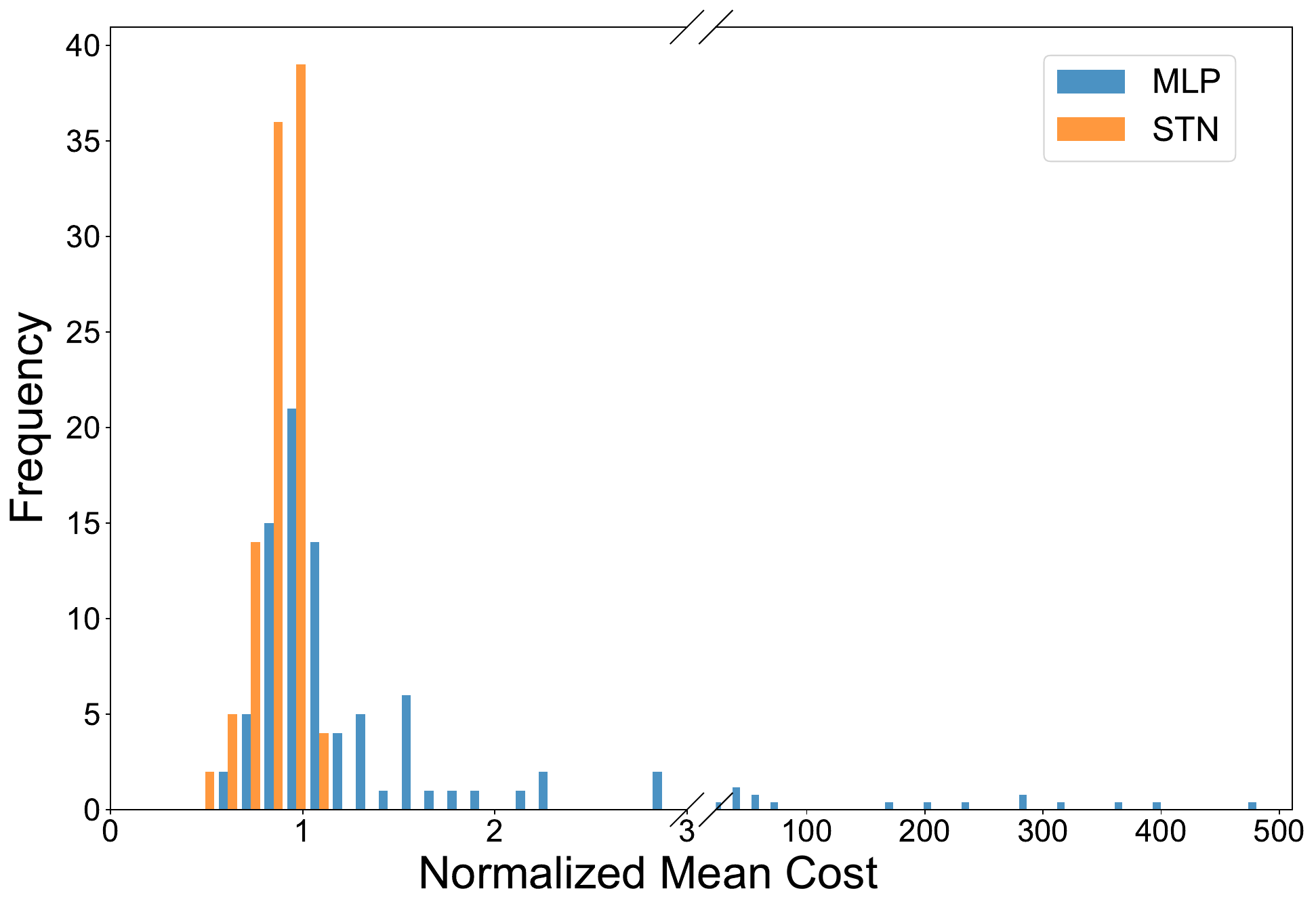}
    \end{subfigure}
    \begin{subfigure}{0.4\textwidth}
        \includegraphics[width=\textwidth]{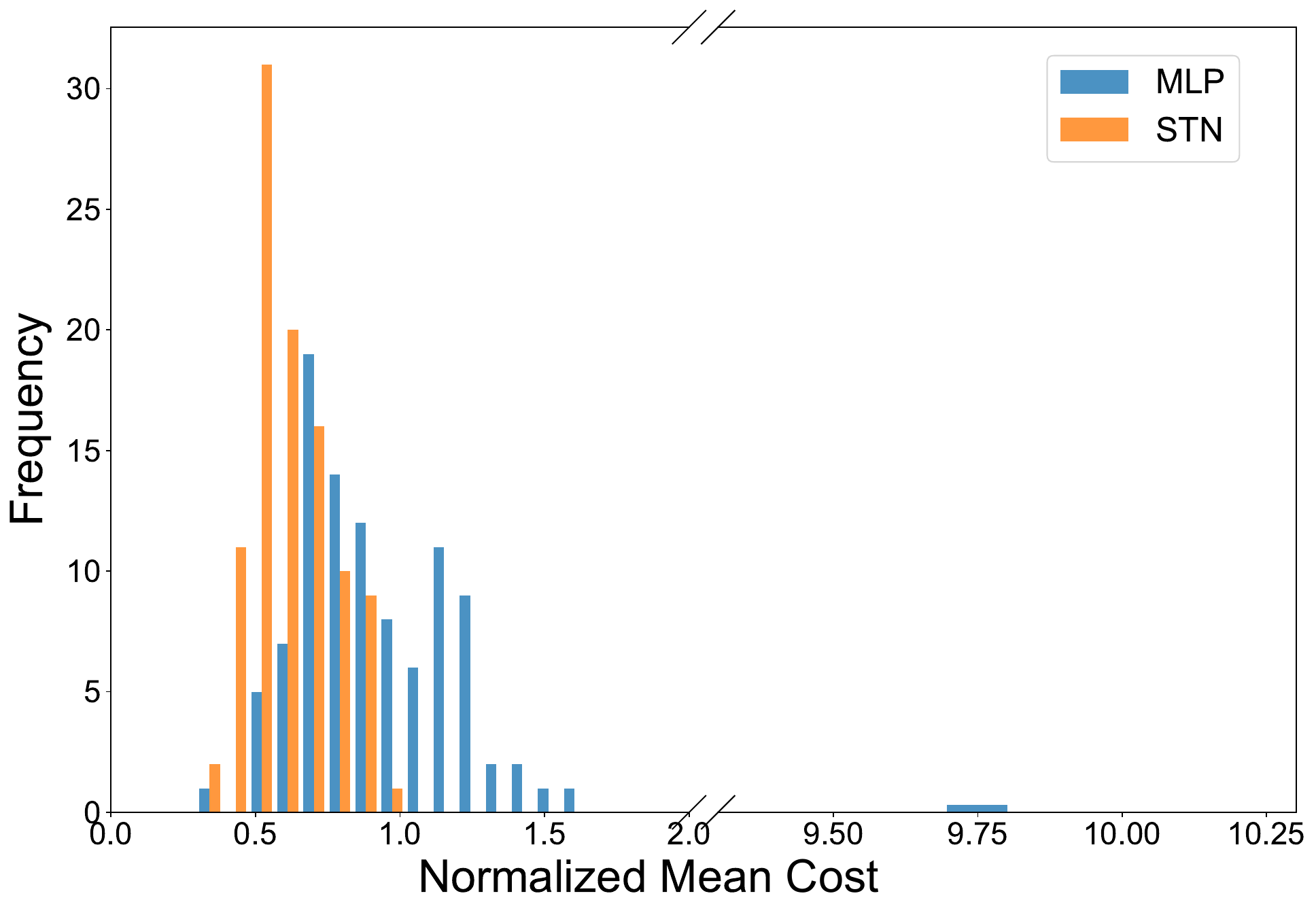}
    \end{subfigure}
    \caption{Histogram of $\hat J_{0}(\pi_{\theta}, \mathcal E^{(j)})$ for the test environments $\mathcal E^{(j)}\in \mathbfcal E_{test}$ for the single-hop scheduling environments (top) and multi-path routing environments (bottom) .  Note the ``break" on the x-axis halfway through the plot, where the scaling changes, where the right half of the plot shows the outliers for each test set.} 
    \label{fig:sp_hist}
\end{figure}

Next we provide the results from the evaluation phase over all environments in $\mathbfcal E$ for both policy network architectures and both environment types.  Once again, we utilize a normalized average cost, but we focus on the averages over the training and testing environment sets. We use the notation $J_0(\pi_{\theta}, \mathbfcal E)$ to denote the average normalized cost over all environments $\mathcal E^{(j)}\in \mathbfcal E$.   For the MLP policy network, on some environments, $J_0(\pi_{\theta, MLP}, \mathcal E^{(j)})$ was very large meaning $\pi_{\theta, MLP}$ was unable to stabilize environment $\mathcal E^{(j)}$ and thus the queue sizes and cost grew without bounds during the evaluation phase.  Thus we also report an outlier rejected average $J_0(\pi_{\theta,MLP}, \mathbfcal{\tilde E})$ which omits any $J(\pi_{\theta, MLP}, \mathcal E^{(j)})$ from the averaging process if $J(\pi_{\theta, MLP}, \mathcal E^{(j)})>5$.  These metrics are reported in Table \ref{tab:SH_tab} for the single-hop environments and Table \ref{tab:MP_tab} for the multi-path environments.  Additionally, histograms of the average cost per test environment $J(\pi_{\theta}, \mathcal E^{(j)})$ for $\mathcal E^{(j)}\in \mathbfcal E_{test}$ for both policy networks and both environment types are shown in \Cref{fig:sp_hist}.  From these tables and figures, we can see that the STN policy network outperforms the MLP policy network on both environments and for both the training and test contexts.  The performance difference is more pronounced for the single-hop environments, which corresponded to a more challenging set of environments as both arrival and service rates varied resulting in more variation between environments. Although these metrics are skewed by the environments in which the MLP policy did not stabilize, the STN network outperformed the MLP policy network by a factor of $8$ and $38.7$ for the training and test single-hop scheduling environments respectively.  However, even when considering the outlier rejected averages, the STN policy network outperforms the MLP policy network on both the training and test single-hop scheduling environments. Additionally, computing $J(\pi_{\theta,MLP}, \mathbfcal {\tilde E}_{train})$ and $\hat J(\pi_{\theta, MLP}, \mathbfcal{\tilde E}_{test})$, 1 context was omitted from $\mathbfcal E_{train}$ and $19$ contexts were omitted from $\mathbfcal E_{test}$.  Meaning at  $20\%$ of the contexts were ignored from the training and test contexts for these metrics for the MLP policy network. The performance difference is less pronounced on the multi-path routing environments, however, the STN policy network does outperform the MLP policy network with regards to every metric. Not only does the STN policy network improve upon the MLP policy network, on-average the STN policy network outperforms the MaxWeight scheduling and Shortest Queue routing algorithms which where used as $\pi_0$.  
\begin{table}
    \centering
    \begin{tabular}{|c|c|c|} \hline 
         $J_0(\pi_{\theta}, \mathbfcal E)$&  MLP& STN\\ \hline 
         $\mathbfcal E_{train}$&  $6.90 \quad (13.39)$& $0.860 \quad (0.049)$\\ \hline 
 $\mathbfcal{\tilde{E}}_{train}$& $0.907 \quad (0.521)$&$0.860 \quad (0.049)$\\ \hline 
         $\mathbfcal E_{test}$&  $33.7 \quad (94.1)$& $0.870 \quad (0.128)$\\ \hline 
         $ \mathbfcal{ \tilde{E}}_{test}$&  $1.20 \quad (0.521)$& $0.870 \quad (0.128)$\\ \hline
    \end{tabular}
    \caption{Mean and standard deviation (in parenthesis) of the normalized average cost during evaluation for both architectures across each set of single-hop environments. For the outlier rejected metrics for the MLP policy network, $1$ context was omitted from $\mathbfcal{ \tilde E}_{train}$ and $19$ contexts were omitted from $\mathbfcal{ \tilde E}_{test}$.} 
    \label{tab:SH_tab}
\end{table}

\begin{table}
    \centering
    \begin{tabular}{|c|c|c|} \hline 
         $J_0(\pi_{\theta}, \mathbfcal E)$&  MLP& STN\\ \hline 
         $\mathbfcal E_{train}$&  $0.571 \quad (0.162)$& $0.550 \quad (0.159)$\\ \hline 
         $\mathbfcal E_{test}$&  $1.03 \quad (0.956)$& $0.625 \quad (0.140)$\\ \hline 
         $ \mathbfcal{ \tilde E}_{test}$&  $0.938 \quad (0.274)$& $0.625 \quad (0.140)$ \\ \hline\end{tabular}
    \caption{ Mean and standard deviation (in parenthesis) of the normalized average cost during evaluation for both architectures across each set of multi-path environments. For the outlier rejected metrics for the MLP policy network, $1$ context was omitted from $\mathbfcal{ \tilde E}_{test}$. }
    \label{tab:MP_tab}
\end{table}

\subsection{Multi-Environment Training Sample Efficiency}
\begin{figure}
    \centering
    \begin{subfigure}[b]{0.4\textwidth}
        \includegraphics[width=\textwidth]{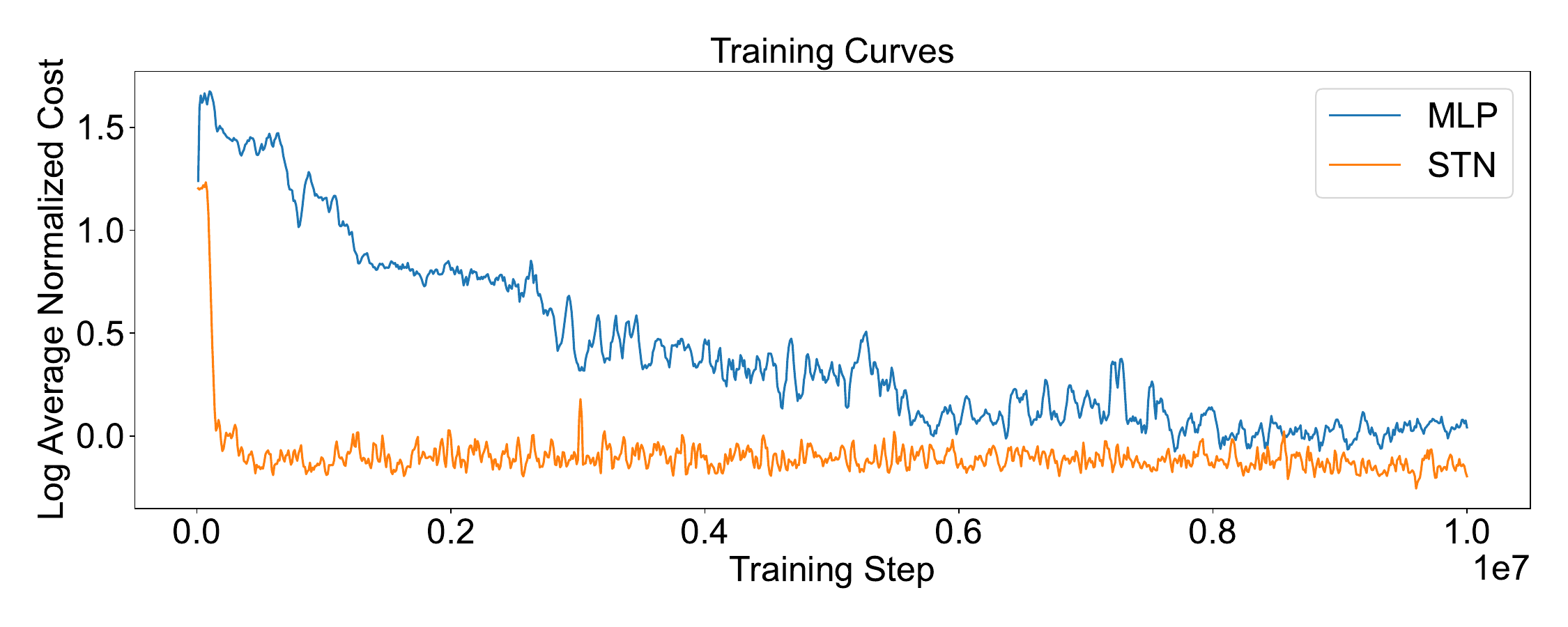}
    \end{subfigure}
    \begin{subfigure}[b]{0.4\textwidth}
        \includegraphics[width=\textwidth]{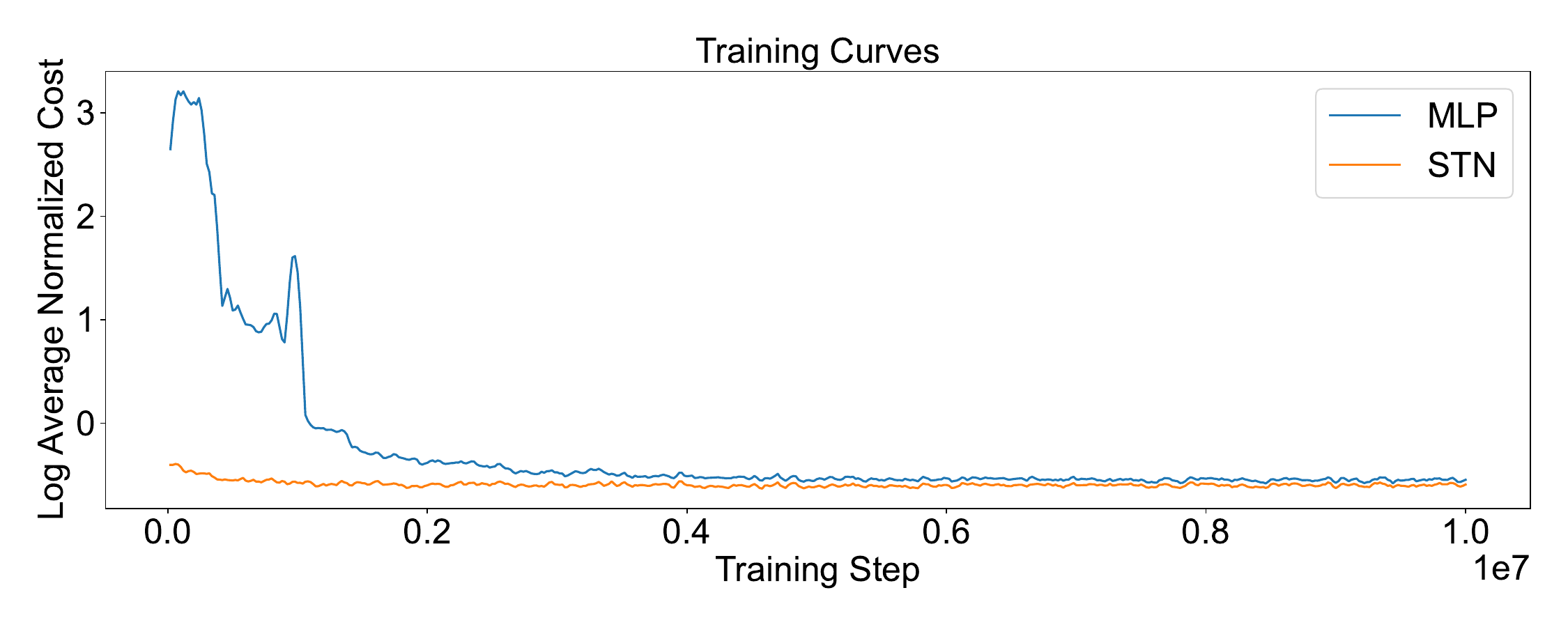}
    \end{subfigure}
    \caption{Moving average cost of the training policy averaged across all five training environments vs the 
training step. Top plot corresponds with the single-hop training environments and the bottom corresponds with the multi-path training environments}
    \label{fig:multi_tr_curve}
\end{figure}

We also compare the training sample efficiency of each architecture.  As in the single environment training procedure, we track the $T_{MA}=5,000$ step moving average cost during training for each environment $\mathcal E^{(j)}\in \mathbfcal E_{train}$, and normalize this metrics by $J(\pi_0, \mathcal E^{(j)})$. We then take an average of this normalized moving average cost across all training environments.  We plot these training curves in \Cref{fig:multi_tr_curve} using the logarithm of the mean normalized cost on the y-axis, as the magnitude of this metric varied widely during training especially for the MLP policy. As seen for the case when a single policy network is trained on a single environment, the STN policy network is significantly more sample efficient than the MLP policy network when training on multiple environments simultaneously.

\section{Conclusion}
In this work, we introduce a novel policy architecture for learning resource allocation control policies: the switch-type policy network (STN). This architecture incorporates domain-specific knowledge derived from a long history of switch-type policies in resource allocation. We show that the STN policy network enhances the sample efficiency of the PPO algorithm without compromising evaluation performance, suggesting that switch-type policies are optimal for the problem classes studied, consistent with similar models in the literature. The STN architecture significantly outperforms the MLP architecture in zero-shot generalization. Unlike the MLP, which tends to overfit to training environments and performs well only when training and testing occur in the same environment, the STN network not only matches the MLP’s performance in familiar environments but also generalizes effectively to unseen ones. This indicates that the optimal policy is switch-type with respect to both queue sizes and arrival/service rate parameters. Future research will focus on demonstrating the optimality of the switch-type policy class for individual and large sets of similar environments.

\bibliographystyle{IEEEtran}
\bibliography{references}

\end{document}